\documentclass{article}
\usepackage{}
\usepackage{amssymb}

     \usepackage[final]{neurips_2019}

\usepackage[utf8]{inputenc} 
\usepackage[T1]{fontenc}    
\usepackage{hyperref}       
\usepackage{url}            
\usepackage{booktabs}       
\usepackage{amsfonts}       
\usepackage{nicefrac}       
\usepackage{microtype}      

  \usepackage{graphicx}
  \usepackage{epstopdf}

\title{Quantization Loss Re-Learning Method}

\author{%
  Kunping Li\thanks{Master postgraduate of Chongqing University, engaged in Natural Language Processing research.} \\
  Chongqing University\\
  Chongqing, China, 400044 \\
  \texttt{likunping@cqu.edu.cn} \\
}

\begin{document}

\maketitle

\begin{abstract}
  In order to quantize the gate parameters of the LSTM (Long Short-Term Memory) neural network model with almost no recognition performance degraded, a new quantization method named Quantization Loss Re-Learn Method is proposed in this paper. The method does lossy quantization on gate parameters during training iterations, and the weight parameters learn to offset the loss of gate parameters quantization by adjusting the gradient in back propagation during weight parameters optimization. We proved the effectiveness of this method through theoretical derivation and experiments. The gate parameters had been quantized to 0, 0.5, 1 three values, and on the Named Entity Recognition dataset, the F1 score of the model with the new quantization method on gate parameters decreased by only 0.7\% compared to the baseline model.
\end{abstract}

\section{Introduction}

  With the rapid development of Artificial Intelligence, the application scenarios of Deep Neural Network are becoming more and more multiplex. Correspondingly, the scale of neural networks is rapidly expanding. How to use a limited computing resource to implement a more complex DNN model, or how to achieve greater data processing throughput under the same computing resource overhead, has become an increasingly important and significant topic in the research of DNN models.

  In order to reduce the actual consumption of the neural network resources on the embedded device and the servers, and to explore the performance effect of quantization on gate parameters of Long Short-Term Memory neural network, a series of experiments were implemented based on the Bi-LSTM (Long Short-Term Memory) model.

  In this paper, we propose a method called Quantization Loss Re-Learning. Specifically, this new method is proposed based on the new idea that the quantization loss of one parameter can be offset by another parameter, i.e. we can use the weight parameter to learn to offset the quantization loss of the gate parameters. In the experiments of this paper, the model implemented by our new method has good performance. By using Gumbel Softmax method proposed in previous work and the new method proposed in this paper, the model design and experiments quantization on gate parameters of LSTM gate parameters are completed.

  In the quantization experiments of gate parameters, this paper selects the 0, 0.5, 1 quantization scheme that can be efficiently implemented in hardware design. After the gate parameters is quantized, a large number of multiplication calculations in the LSTM model can be converted into addition and shift operations. On the Named Entity Recognition dataset, the F1 score of the model with the new quantization method on gate parameters decreased by only 0.7\% compared to the baseline model.

  Limitation of computational resource affects the application of DNN models. Therefore, it is of great practical significance to research the method on reducing computational resource dependence of DNN. In Addition, although our research focuses on the LSTM model, the research on Quantitative Loss Migration Learning or Quantitative Loss Re-Learning is still at a very early stage. Our research can be used as a very early and important case in this field.

In the previous research [14], by forcing the output value of the gates to 0 or 1, that mean the gates are fully open or closed, the gates of the LSTM model can better control the information flow [5]. This makes the gates of the LSTM model more interpretable. And previous research shows that [5]: pushing the gate parameters close to 0 or 1 will intuitively cause performance loss to the LSTM model, but in fact this does not cause significant performance loss, in some case, the generalization ability of the LSTM model can even be improved.

\section{Related Work}

In the previous research, by forcing the output value of the gates to 0 or 1, that mean the gates are fully open or closed, the gates of the LSTM model can better control the information flow. This makes the gates of the LSTM model more interpretable. And previous research shows that: pushing the gate parameters close to 0 or 1 will intuitively cause performance loss to the LSTM model, but in fact this does not cause significant performance loss, in some case, the generalization ability of the LSTM model can even be improved.

\subsection{Gumbel Softmax}

The Gumbel distribution, which named after the mathematician Emil Julius Gumbel, is an extreme distribution, also known as a double exponential distribution. In probability and statistics theory, the Gumbel distribution is used to simulate the maximum or minimum distribution of multiple samples of various distributions. By using the Gumbel Softmax method, the random variable $R_{\mathrm{U}}$, which originally obeys the standard normal distribution, is transformed into a random variable $R_{\mathrm{B}}$, which obeys the discrete distribution. According to this method, the paper first proposes to approximately replace the binomial distribution with the Gumbel distribution, where the function $\mathrm{G}\left(\alpha ,\varepsilon \right)$ is defined as Equation(3).

In the research of the paper [5], only the output values of the input gate and the forget gate are forced to approach 0 or 1, without applying similar operations on the output value of the output gate, because the output gate usually requires more fine-grained information for decision making, and therefore output gate is not suitable for quantization.
\begin{equation}
\mathrm{P}\left(\mathop{\mathrm{lim}}_{\varepsilon \mathrm{\to }0^{\mathrm{+}}}\mathrm{G}\left(\alpha ,\varepsilon \right)\mathrm{=1}\right)\mathrm{=}\mathrm{P}\left(R_{\mathrm{B}}\mathrm{=1}\right)
\end{equation}
\begin{equation}
\mathrm{P}\left(\mathop{\mathrm{lim}}_{\varepsilon \mathrm{\to }0^{\mathrm{+}}}\mathrm{G}\left(\alpha ,\varepsilon \right)\mathrm{=0}\right)\mathrm{=}\mathrm{P}\left(R_{\mathrm{B}}\mathrm{=0}\right)
\end{equation}
\begin{equation}
\mathrm{G}\left(\alpha ,\varepsilon \right)\mathrm{=}\sigma \left(\frac{\alpha \mathrm{+}{\mathrm{log} R_{\mathrm{U}}\ }\mathrm{-}{\mathrm{log} \left(\mathrm{1-}R_{\mathrm{U}}\right)\ }}{\varepsilon }\right)
\end{equation}

In forward propagation, each iteration of the training needs to independently sample from Normal Distribution, and then use the Equation(1) and Equation(2) to calculate the gate parameters by using the Gumbel Softmax method. $R_{\mathrm{G}}$ is continuous and differentiable with respect to parameters of LSTM, and the losses are continuous and differentiable with respect to $R_{\mathrm{G}}$, so any standard gradient based method can be used to update the parameters [5]. But obviously Gumbel Softmax method is too complicated to quantize gate parameters.

\section{Quantization Loss Re-learn Method}

\subsection{Theorem}

U, V, and W respectively correspond to the weight matrix of the connection from the input unit to the cell unit, the cell unit to the hidden unit, and the hidden unit to the cell unit in the LSTM model [8]. Where
{U}={U}\textit{${}_{C}$}$\mathrm{\cup}${U}\textit{${}_{i}$}$\mathrm{\cup}${U}\textit{${}_{f}$}$\mathrm{\cup}${U}\textit{${}_{o}$}, {V}={V}\textit{${}_{C}$}$\mathrm{\cup}${V}\textit{${}_{i}$}$\mathrm{\cup}${V}\textit{${}_{f}$}$\mathrm{\cup}${V}\textit{${}_{o}$}, {W}={W}\textit{${}_{C}$}$\mathrm{\cup}${W}\textit{${}_{i}$}$\mathrm{\cup}${W}\textit{${}_{f}$}$\mathrm{\cup}${W}\textit{${}_{o}$} Subscript \textit{C}, \textit{i}, \textit{f}, \textit{o} represent cell state, input gate, forget gate, output gate respectively. The state transfer formula of the LSTM model is equivalently rewritten as Equation(4-9):

\begin{equation}
i^{\left(\mathrm{t}\right)}_{\ }\mathrm{=}\sigma \left({{\mathrm{U}}}_ix^{\left(\mathrm{t}\right)}\mathrm{+}{{\mathrm{W}}}_ih^{\left(\mathrm{t-1}\right)}\mathrm{+}{{\mathrm{V}}}_iC^{\left(\mathrm{t-1}\right)}\mathrm{+}b_i\right)
\end{equation}
\begin{equation}
f^{\left(\mathrm{t}\right)}_{\ }\mathrm{=}\sigma \left({{\mathrm{U}}}_fx^{\left(\mathrm{t}\right)}\mathrm{+}{{\mathrm{W}}}_fh^{\left(\mathrm{t-1}\right)}\mathrm{+}{{\mathrm{V}}}_fC^{\left(\mathrm{t-1}\right)}\mathrm{+}b_f\right)
\end{equation}
\begin{equation}
C^{\left(\mathrm{t}\right)}\mathrm{=}f^{\left(\mathrm{t}\right)}_{\ }C^{\left(\mathrm{t-1}\right)}\mathrm{+}i^{\left(\mathrm{t}\right)}_{\ }\mathrm{tanh}\left({{\mathrm{U}}}_Cx^{\left(\mathrm{t}\right)}\mathrm{+}{{\mathrm{W}}}_Ch^{\left(\mathrm{t-1}\right)}\mathrm{+}b_C\right)
\end{equation}
\begin{equation}
o^{\left(\mathrm{t}\right)}_{\ }\mathrm{=}\sigma \left({{\mathrm{U}}}_ox^{\left(\mathrm{t}\right)}\mathrm{+}{{\mathrm{W}}}_oh^{\left(\mathrm{t-1}\right)}\mathrm{+}{{\mathrm{V}}}_oC^{\left(\mathrm{t}\right)}\mathrm{+}b_o\right)
\end{equation}
\begin{equation}
h^{\left(\mathrm{t}\right)}\mathrm{=}o^{\left(\mathrm{t}\right)}_{\ }\mathrm{tanh}\left({{\mathrm{V}}}_CC^{\left(\mathrm{t}\right)}\right)
\end{equation}
\begin{equation}
{\hat{y}}^{\left(\mathrm{t}\right)}\mathrm{=softmax}\left(h^{\left(\mathrm{t}\right)}\right)
\end{equation}

%

In order to analyze how the quantization of the gate parameters affects the performance of the LSTM, the gradient formula of the backpropagation process needs to be derived. Based on the LSTM forward propagation calculation process, the LSTM backpropagation algorithm analytical formula is derived. According to the derivation of previous research, the gradient formula is defined as Equation(10).
\begin{equation}
\frac{\partial L}{\partial {{\mathrm{W}}}_f}\mathrm{=}\sum^{\tau }_{\mathrm{t=1}}{\frac{\partial L}{\partial C^{\left(\mathrm{t}\right)}}\frac{\partial C^{\left(\mathrm{t}\right)}}{\partial f^{\left(\mathrm{t}\right)}}\frac{\partial f^{\left(\mathrm{t}\right)}}{\partial {{\mathrm{W}}}_f}}\mathrm{=\ }\sum^{\tau }_{\mathrm{t=1}}{\left[{\delta }^{\left(\mathrm{t}\right)}_C\mathrm{\odot }C^{\left(\mathrm{t-1}\right)}\mathrm{\odot }f^{\left(\mathrm{t}\right)}\mathrm{\odot }\left(\mathrm{1-}f^{\left(\mathrm{t}\right)}\right)\right]{\left(h^{\left(\mathrm{t-1}\right)}\right)}^{\mathrm{T}}}
\end{equation}

We suppose that $f^{\left(\mathrm{t}\right)}={\overline{f}}^{\ \left(\mathrm{t}\right)}+\Delta f^{\left(\mathrm{t}\right)}$, where ${\overline{f}}^{\ \left(\mathrm{t}\right)}$ is the result of the approximate fixed point value after the Round \& Clip operation, And $\Delta f^{\left(\mathrm{t}\right)}$represents the difference between the approximate result and the exact result. The vector multiplication of $f^{\left(\mathrm{t}\right)}\mathrm{\odot }\left(\mathrm{1-}f^{\left(\mathrm{t}\right)}\right)$ is expanded and rearranged like Equation(11).

\begin{equation}
\left({\overline{f}}^{\mathrm{\ }\left(\mathrm{t}\right)}\mathrm{+}\mathrm{\Delta }f^{\left(\mathrm{t}\right)}\right)\mathrm{\odot }\left(\mathrm{1-}{\overline{f}}^{\mathrm{\ }\left(\mathrm{t}\right)}\mathrm{-}\mathrm{\Delta }f^{\left(\mathrm{t}\right)}\right)={\overline{f}}^{\mathrm{\ }\left(\mathrm{t}\right)}\mathrm{\odot }\left(\mathrm{1-}{\overline{f}}^{\mathrm{\ }\left(\mathrm{t}\right)}\right)\mathrm{+}\mathrm{\Delta }f^{\left(\mathrm{t}\right)}\mathrm{\odot }\mathrm{(}\mathrm{1-2}{\overline{f}}^{\mathrm{\ }\left(\mathrm{t}\right)}\mathrm{-}\mathrm{\Delta }f^{\left(\mathrm{t}\right)}\mathrm{)}
\end{equation}

Then, as Equation(12),
\[\frac{\partial L}{\partial {{W}}_f}=\sum^{\tau }_{t=1}{\left[{\delta }^{\left(t\right)}_C\odot C^{\left(t-1\right)}\odot {\overline{f}}^{\ \left(t\right)}\odot \left(1-{\overline{f}}^{\ \left(t\right)}\right)\right]{\left(h^{\left(t-1\right)}\right)}^T}\]
\begin{equation}
+\sum^{\tau }_{t=1}{\left[{\delta }^{\left(t\right)}_C\odot C^{\left(t-1\right)}\Delta f^{\left(t\right)}\odot (1-2{\overline{f}}^{\ \left(t\right)}-\Delta f^{\left(t\right)})\right]{\left(h^{\left(t-1\right)}\right)}^T}
\end{equation}

Where $\mathrm{-}\mathrm{1}\mathrm{\le }\mathrm{\forall }\mathrm{\ }\mathrm{ele}\mathrm{ments}\mathrm{\in }\left(1-2{\overline{f}}^{\ \left(\mathrm{t}\right)}-\Delta f^{\left(\mathrm{t}\right)}\right)\mathrm{\le }\mathrm{+1}$. So the second term is apparent approaches 0 when $\Delta f^{\left(\mathrm{t}\right)}\to 0$. and the back propagation gradient of the first term is equivalent to the gradient formula before quantization. This indicates that during the optimization process, ${{\mathrm{W}}}_f$ will be optimized toward the direction in which $\Delta f^{\left(\mathrm{t}\right)}$ is reduced, and finally ${{\mathrm{W}}}_f$ learns to offset the loss caused by the Round \& Clip quantization operation.

Further, from $f^{\left(\mathrm{t}\right)}\mathrm{=}\sigma \left({{\mathrm{W}}}_fh^{\left(\mathrm{t-1}\right)}\mathrm{+}{{\mathrm{U}}}_fx^{\left(\mathrm{t}\right)}\mathrm{+}b_f\right)$, it can be known that the generation of$f^{\left(\mathrm{t}\right)}$ is determined by ${{\mathrm{W}}}_f$ and ${{\mathrm{U}}}_f$ together, the backpropagation gradient of ${{\mathrm{U}}}_f$ is similar to the gradient of ${{\mathrm{W}}}_f$, so the loss generated by quantization will be learned and offset by ${{\mathrm{W}}}_f$ and ${{\mathrm{U}}}_f$. The quantization loss of other parameters is offset by a similar process.

\subsection{Models Design}

\begin{figure}
  \centering
  \includegraphics[scale=0.9]{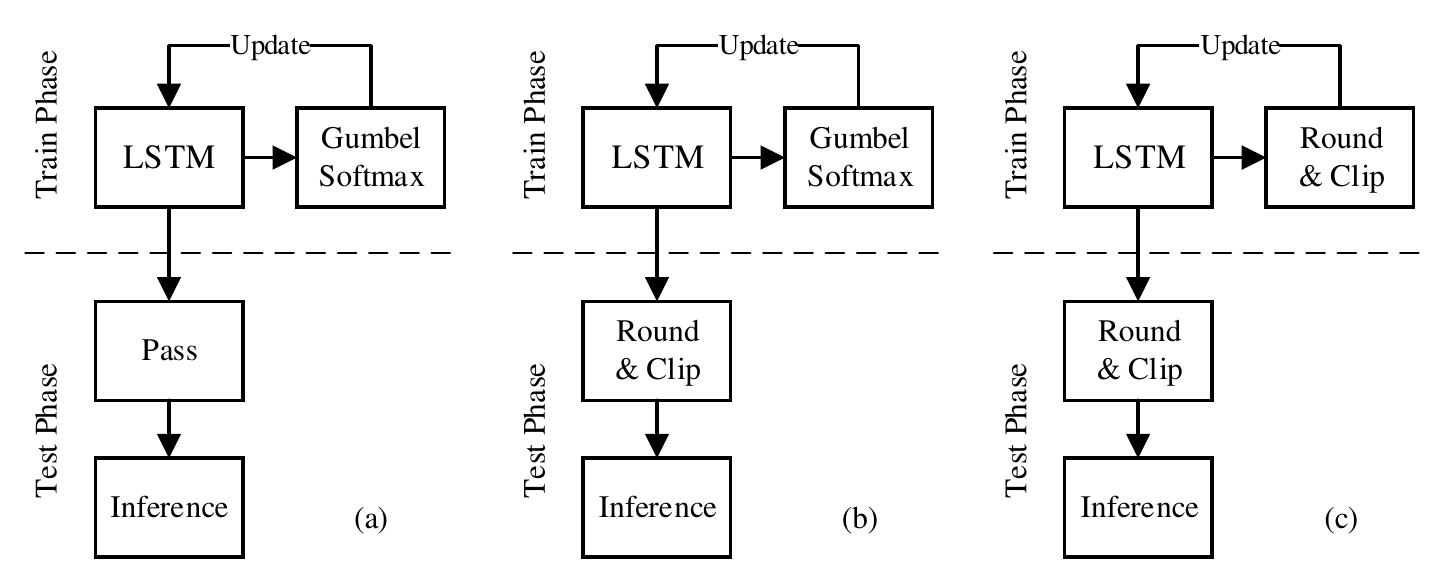}

  \caption{Gumbel Softmax and Round \& Clip working positions: (a) Gumbel Softmax works during the training phase singly; (b) Gumbel Softmax works during the training phase, then Round \& Clip quantization is used; (c) Round \& Clip works during the training phase, then Round \& Clip quantization is used.}
\end{figure}

Figure 1 (a) and (b) show the model calculation graph that using the Gumbel Softmax method to force the gate parameters to approach 0 or 1. The LSTM calculation box in the figure represents the set of all parameters in the LSTM model, including weight parameters and gate parameters. In these figures, the Gumbel Softmax calculation box and the Round \& Clip calculation box are only used for the gate parameters.

Figure 1 (a) shows that after the training with the Gumbel Softmax method, the Round \& Clip method is not used for post-training quantization. Figure 4.1 (b) show that after the training with the Gumbel Softmax method and then using the Round \& Clip method for post-training quantization.

Figure 1 (c) shows the calculation graph of the new gate parameters quantization method proposed in this paper. In the calculation graph of the new method, the original Gumbel Softmax calculation box is replaced by the Round \& Clip calculation box. That means each iteration in the training uses Round \& Clip to quantize the gate parameters and then update the LSTM parameters. In the testing phase, the LSTM gate parameters will be directly quantized by Round \& Clip. The calculated path is LSTM, Round \& Clip, and Inference. The meaning of Inference is that when the LSTM determines the parameter space from the input sequence information, the output sequence is inferred from the input sequence.

\section{Experiments}

In the experiment section of this paper, we will validate and explore the most representative results in the field of LSTM gate parameters quantization in recent years. Inspired by those results, we propose a new algorithm that introduces the new gate parameters quantization method named Quantization Loss Re-Learning. By using this new method, the gate parameters of LSTM can be quantized closer to a few fixed values without significantly degrading performance.

\subsection{Setup}

\textbf{Experimental data: } The three datasets CoNLL2000\_chunking, GermEval 2014 and Unidep\_POS is used for experiments in this section. The size of the dataset is shown in Table 1, where the numerical value indicates the number of samples.

\begin{table}
  \caption{Dataset setting}
  \label{sample-table}
  \centering
  \begin{tabular}{cccccc}
    \toprule

    No. & Task & Dataset & Train-Phase & Dev-Phase & Test-Phase \\

    \midrule
1 & NER & GermEval 2014 & 24000 & 12543 & 8936\\
2 & Chunking & CoNLL2000\_chunking&5100&2007&2012 \\
3 & POS & Unidep\_POS&2200&2002&1844 \\
    \bottomrule
  \end{tabular}
\end{table}

\textbf{Control settings: } The experiments described in this section mainly include the following sub-experiments. Performance changes after the gate parameters quantization using Gumbel Softmax. Performance changes after the gate parameters quantization using the Round \& Clip. Performance changes after gate parameters quantization using the new method.

\textbf{Method of processing the experimental results: }In the Settings field of tables, the initial letter G indicates the Gumbel Softmax method, and B indicates quantization operation. The last letter I indicates the input gate, and F indicates the forget gate. For example, BI represents an input gate that has been processed using a quantization method.

\subsection{Gumbel Softmax}

The purpose of this section is to verify the effectiveness of the Gumbel Softmax method. The experiments will compare the performance variation of the model before and after using Gumbel Softmax method. We have respectively experimented on the following gate structures: the input gate I, the forget gate F, and the output gate O. And F1 Score and other performance indicators are tested and recorded. In addition, in the last experiment of this section, according to the experiment result in paper [5] that doing Gumbel Softmax on the output gate results in performance degradation, the Gumbel Softmax on output gate is ignored similarly in our experiment, and the last experiment in this section combines only GI and GF operations.

From Table 2, we can see the change of the average F1 Score before and after using Gumbel Softmax. The benchmark model is Bi-LSTM-CRF, while the model using Gumbel Softmax named GI, GF and so on. The F1 Score difference between the model using GI and GF and the benchmark model is only 0.002, which means almost no performance loss by using Gumbel Softmax. The difference between the GO model and the benchmark model is 0.016, and this difference is obvious. After the combine of GI and GF, the performance of the model increased slightly by 0.003. This phenomenon is completely consistent with the description in the paper [5]. That is, since the value of the gate parameters tends to 0 or 1, the gate parameters will become more meaningful, that 1 means pass and 0 means close, and thus the generalization ability of the model can be promoted. The experimental results on more datasets are shown in Table 3.

This experiment corresponds to the model design shown in Figure 1 (a). The simple use of Gumbel Softmax has little effect on F1 Score, but it will additionally increase the random number generation process. And although performance has not declined by using Gumbel Softmax simply, storage resources also have no declined.

\begin{table}
  \caption{Performance impact of Gumbel Softmax on gate parameters of LSTM (NER)}
  \label{sample-table}
  \centering
  \begin{tabular}{cccccccc}
    \toprule

    No. & Settings & Epoch & Time\_diff & Time\_total & Prec & Rec & F1 \\
    \midrule

1 & Bi-LSTM-CRF & 24.7 & 109.7 & 2706 & 0.814 & 0.779 & 0.788 \\
2 & GI & 24.0 & 114.8 & 2755 & 0.817 & 0.777 & 0.786 \\
3 & GF & 22.0 & 115.7 & 2545 & 0.820 & 0.783 & 0.789 \\
4 & GO & 25.0 & 119.8 & 2994 & 0.807 & 0.749 & 0.772 \\
5 & GI, GF & 20.3 & 122.7 & 2497 & 0.822 & 0.780 & 0.791 \\
    \bottomrule
  \end{tabular}
\end{table}

\noindent

\begin{table}
  \caption{Performance impact of Gumbel Softmax on gate parameters of LSTM}
  \label{sample-table}
  \centering
  \begin{tabular}{cccccccc}
    \toprule

    No. & Settings & Epoch & Time\_diff & Time\_total & Prec & Rec & F1 \\
    \midrule

1 & GermEval 2014 & 20.3 & 122.7 & 2497 & 0.822 & 0.78 & 0.791 \\
2 & CoNLL2000\_chunking & 25.0 & 76.7 & 1917 & 0.941 & 0.938 & 0.939 \\
    \bottomrule
  \end{tabular}
\end{table}

\subsection{Round \& Clip}

This experiment corresponds to the model design shown in Figure 1 (b). When the gate parameters tend to 0 or 1, the theoretical performance degradation after the Round \& Clip operation will be small. In order to verify that the gate parameters using Gumbel Softmax method is approaching 0 or 1, we perform a Round \& Clip operation to previous models in this section. We use the Round \& Clip quantization method to quantize floating-point gate parameters with values between the intervals (0, 1) to three fixed values.

In this process, round(x) = round(x / r) * r, and clip(x) = clip(x, -c, c). When using the hyperparameter setting of r=0.5 and c=1.0, the result will be quantized into three fixed point values of 0, 0.5, and 1. This quantization scheme is very beneficial for efficient neural network hardware design. The gate parameters are matrix multiplication with other parameters inside the LSTM unit, where the gate parameters are vector. When the quantization result is 0, 0.5, 1.0 three fixed point values, in the case of hardware implementation, it is easy to use hardware operations such as clear, right shift, and accumulated to implement.

As shown in Table 4, when the model has parameters r=0.2 and c=0.4, the performance is better than r=0.5 and c=1.0. In the GermEval 2014 data set, when using the 0, 0.5, and 1.0 quantization schemes, The F1 Score is only a little higher than half of the F1 Score when using the 0, 0.2, and 0.4 quantization schemes. More experimental results on multiple datasets is shown in Table 5.

This indicates that when the Round \& Clip operation is simply performed, the storage resource consumption can be reduced, but there is no guarantee that the quantization scheme can be easily implemented in hardware. And after using Round \& Clip operation, the performance degradation is obvious, although the Gumbel Softmax method is used in training, it has little effect.

\begin{table}
  \caption{Performance impact of Round \& Clip on gate parameters of LSTM (NER)}
  \label{sample-table}
  \centering
  \begin{tabular}{ccccccccc}
    \toprule
    No. & Settings & r / c & Epoch & Time\_diff & Time\_total & Prec & Rec & F1 \\

    \midrule
1 & Bi-LSTM-CRF & - / - & 24.7 & 109.7 & 2706 & 0.814 & 0.779 & 0.788 \\
2 & GI, GF, BI, BF & 0.5 / 1.0 & 23.7 & 251.3 & 6005 & 0.818 & 0.23 & 0.358 \\
3 & GI, GF, BI, BF & 0.2 / 0.4 & 25.0 & 89.0 & 2224 & 0.844 & 0.587 & 0.688 \\
    \bottomrule
  \end{tabular}
\end{table}

\begin{table}
  \caption{Performance impact of Round \& Clip on gate parameters of LSTM}
  \label{sample-table}
  \centering
  \begin{tabular}{ccccccccc}
    \toprule
    No. & Settings & r / c & Epoch & Time\_diff & Time\_total & Prec & Rec & F1 \\

    \midrule
1 & NER & 0.5 / 1.0 & 23.7 & 251.3 & 6005 & 0.818 & 0.23\textbf{} & 0.358\textbf{} \\
2 & Chunking & 0.5 / 1.0 & 19.7 & 80.8 & 1578 & 0.489 & 0.43 & 0.453 \\
3 & POS & 0.5 / 1.0 & 22.3 & 58.3 & 1301 & 0.793\textbf{} & - & - \\
4 & NER & 0.2 / 0.4 & 25.0 & 89.0 & 2224 & 0.844 & 0.587 & 0.688 \\
5 & Chunking & 0.2 / 0.4 & 12.0 & 79.5 & 954 & 0.752 & 0.803 & 0.777 \\
6 & POS & 0.2 / 0.4 & 15.0 & 38.4 & 576 & 0.926 & - & - \\
    \bottomrule
  \end{tabular}
\end{table}

\subsection{New Method}

In order to prove an algorithm is effective, datasets and hyperparameters are carefully selected, this does not mean much. Designing a good model is to hope that this model can learn in Artificial Intelligence way as much as possible, instead of changing a dataset with another try to manually find the best hyperparameters. This is Manual Intelligence.

The Gumbel Softmax method introduces hyperparameters such as r, c, etc. Since Gumbel Softmax is very sensitive to the hyperparameters r, c, it is necessary to pay attention to the setting of these hyperparameter, which brings difficulties to optimize the model [17].


At the same time, the Gumbel Softmax method requires a random number generator to generate Uniform Distribution and Binomial Distribution. It is very expensive to design a high-performance random number generator in hardware design. Therefore, Gumbel Softmax is actually not suitable for implementation on hardware. However, the concept used by Gumbel Softmax is very useful. Experiments by Gumbel Softmax have shown that the distribution of gate parameters can be changed without affecting overall performance.

Considering that Gumbel Softmax adds a random noise to the gate parameters to make the parameters deviates from the true value. By referring to the idea of Gumbel Softmax, this paper proposes a new algorithm for gate parameters quantization, which can degenerate the quantization loss after Round \& Clip operation. 

The previous algorithm is shown like Figure 1 (b). Gumbel Softmax push the gate parameters towards 0 or 1 by adding noise in each iteration, and the Round \& Clip method is used to quantize the gate parameters that tend to 0 or 1 after all training process are completed. This algorithm  has been proved to be unable to achieve good performance.

\begin{table}
  \caption{Performance impact of Degenerate Quantization Loss on gate parameters of LSTM}
  \label{sample-table}
  \centering
  \begin{tabular}{cccccccc}
    \toprule

   No. & Settings & Epoch & Time\_diff & Time\_total & Prec & Rec & F1 \\
    \midrule
1 & Bi-LSTM-CRF & 24.7 & 109.7 & 2706 & 0.814 & 0.779 & 0.788 \\
2 & BI, BF & 25.0 & 110.8 & 3494 & 0.807 & 0.364 & 0.494 \\
3 & BI, BF, B(UVW) & 22.0 & 68.4 & 1504 & 0.796 & 0.488 & 0.598 \\
4 & GI, GF, BI, BF & 23.7 & 251.3 & 6005 & 0.818 & 0.23 & 0.358 \\
5 & BI, BF, NEW & 25.0 & 50.0 & 1251 & 0.821 & 0.761 & 0.781 \\
6 & BI, BF, B(UVW), NEW & 22.7 & 86.5 & 1961 & 0.82 & 0.765 & 0.785 \\
    \bottomrule
  \end{tabular}
\end{table}

\begin{figure}
  \centering
  \includegraphics[scale=0.6]{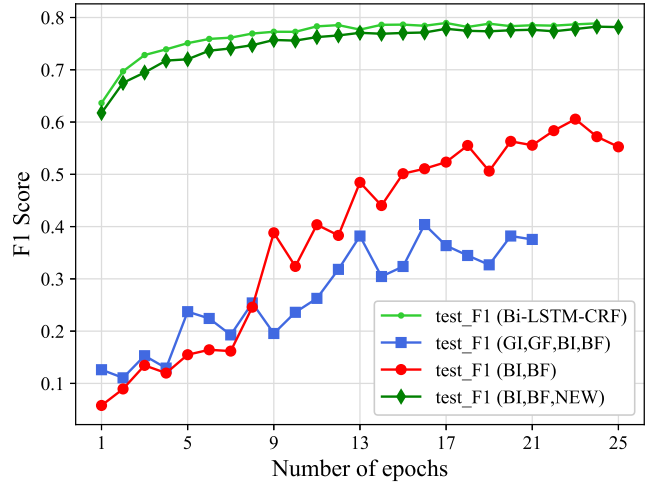}

  \caption{ Performance comparison of Degenerate Quantization Error quantization methods with other quantization methods and benchmark models on gate parameters I, F of LSTM in testing phase.}
\end{figure}

As shown in Figure 1 (c), the new algorithm in this paper uses Round \& Clip to directly quantize the LSTM gate parameters in the current iteration and directly quantize it into the 0, 0.5, and 1.0 schemes. By deriving the gradient formula of the backpropagation of LSTM, it is clear the gate parameters directly affect the gradient used to optimize the weight parameters matrix. Therefore, the loss generated by the gate parameters quantization will eventually be learned and offset by the weight parameters matrix. Therefore, by learning the quantization loss of the gate parameters with the weight matrix, a new and efficient method for quantize the LSTM gate parameters is proposed.

Table 6 and Figure 2 show that if the new algorithm is not used during training, then the gate parameters are directly quantized by Round \& Clip, the performance degradation is very significant. With the NEW method, since the loss of the quantization of the gate parameters is learned and offset by the weight matrix in each iteration, the final experimental results of BI, BF, and NEW are reduced slightly. Compared to the full-precision benchmark model Bi-LSTM-CRF, the model with NEW method decreased by only 0.007, while the BI, BF, B (UVW), NEW model only decreased by 0.003. These results show that the new algorithm proposed in this paper is very effective. Quantization Loss Re-Learning Method can effectively compress the gate parameters, with the performance degradation very slight.

In addition, taking NER as an example, when the quantization scheme is 0, 0.5, and 1.0, all the results are summarized in Figure 2. It can be seen when Round \& Clip is quantized after Gumbel Softmax, the test phase F1 Score is terrible, even worse than using Round \& Clip directly. But the performance of the NEW quantization method proposed in this paper has almost the same performance as the benchmark model. Therefore, it is obvious that Quantization Loss Re-Learning Method is better than using Gumbel Softmax directly, or using Round \& Clip directly, as well as using Gumbel Softmax and Round \& Clip coherently.

\subsection{Conclusion}

This paper focuses on the quantification method of neural network model, and proves that the first-order residual quantization method can be extended by the convolutional neural network to the weight parameter quantization of LSTM, and the performance loss caused by quantization is effectively reduced. Furthermore, by studying the Gumbel Softmax method, this paper proposes a new method of quantifying the gate parameters using the weight parameter to learn the quantization loss of the gate parameters, and proves the effectiveness of this new quantification method through experiments. In the end, this paper combines the weight parameter quantization and the gate parameter quantization to give a set of LSTM efficient quantization scheme.

By analyzing the actual performance of the Gumbel Softmax method in the experiment, this paper draws on the design idea of Gumbel Softmax, and innovatively proposes to quantify the gate parameters in each iteration, and then learn the gate parameters from the weight parameters. An improved idea of quantifying loss and offsetting them. According to this paper, a new LSTM gate parameters quantification method is proposed, and the effectiveness of this method is proved in the corresponding experimental links. That is, when the optimal quantization scheme is quantized, the new gated parameter quantization method F1 Score is reduced by 0.7\% compared to the reference model Bi-LSTM-CRF, and in contrast, the Gumbel Softmax quantization method F1 Score is compared to the reference model. The decrease of 29.4\%, the new gate parameters quantification method proposed in this paper is much better than the Gumbel Softmax quantification method, and has better universality.

In this paper, the 0, 0.5, 1 quantization scheme that can be efficiently implemented in hardware design is selected as the target quantization scheme of the new algorithm. After the above quantization, a large number of multiplication calculations in the LSTM model can be converted into addition and shift operations. The new quantization method proposed in this paper and the quantization result generated by the corresponding algorithm can easily correspond to the corresponding hardware operation in the hardware design. Therefore, based on this algorithm, a set of matching LSTM hardware acceleration design can be further studied. This is the significance of this article and the full text of the study.

\section*{References}

\medskip

\small

[1]	Rastegari, M.,\ Ordonez, V.,\ Redmon, J.,\ \& Farhadi, A.\ (2016). XNOR-Net: ImageNet Classification Using Binary Convolutional Neural Networks. \textit{european conference on computer vision},\ 525-542.

[2]	Courbariaux, M.,\ Hubara, I.,\ Soudry, D.,\ El-Yaniv, R.,\ \& Bengio, Y. (2016). Binarized neural networks: training deep neural networks with weights and activations constrained to +1 or -1. \textit{arXiv}:\textit{ Learning},\ 1602.02830.

[3]	Courbariaux, M.,\ Bengio, Y.,\ \& David, J. P.\ (2015). Binaryconnect: training deep neural networks with binary weights during propagations. \textit{Neural Information Processing Systems},\ 3123-3131.

[4]	Hochreiter, S.,\ \& Schmidhuber, J.\ (1997). Long short-term memory. \textit{Neural Computation,} \textit{9},\ 1735-1780.

[5]	Li, Z., He, D., Tian, F., Chen, W., Qin, T., \& Wang, L., et al. (2018). Towards binary-valued gates for robust LSTM training. \textit{International Conference on Machine Learning},\ 2995-3004.

[6]	Hammerton, J.\ (2003). Named entity recognition with long short-term memory. \textit{Conference on Natural Language Learning at Hlt-naacl}.\ 4: 172-175.

[7]	Mikolov, T.,\ Karafiat, M.,\ Burget, L., Cernock J.,\ \& Khudanpur, S. (2010). Recurrent neural network based language model. \textit{conference of the international speech communication association},\ 1045-1048.

[8]	Bengio, Y.,\ Simard, P. Y.,\ \& Frasconi, P. (1994). Learning long-term dependencies with gradient descent is difficult. \textit{IEEE Transactions on Neural Networks}, 5,\ 157-166.

[9]	Zhang, Y.,\ Chen, G.,\ Yu, D.,\ Yaco, K.,\ Khudanpur, S.,\ \& Glass, J. R.\ (2016). Highway long short-term memory RNNS for distant speech recognition. \textit{international conference on acoustics, speech, and signal processing},\ 5755-5759.

[10]	Sundermeyer, M.,\ Schluter, R.,\ \& Ney, H.\ (2012). LSTM Neural Networks for Language Modeling. \textit{conference of the international speech communication association},\ 194-197.

[11]	Dauphin, Y. N.,\ \& Bengio, Y.\ (2013). Big Neural Networks Waste Capacity. \textit{arXiv: Learning},\ 1301.3583

[12]	Kim, M.,\ \& Smaragdis, P.\ (2016). Bitwise Neural Networks. \textit{arXiv: Learning}, 1601.06071.

[13]	Han, S.,\ Mao, H.,\ \& Dally, W. J.\ (2016). Deep Compression: Compressing Deep Neural Networks with Pruning, Trained Quantization and Huffman Coding. \textit{arXiv: Learning},\  1510.00149

[14]	Cybenko, G.\ (1989). Approximation by superpositions of a sigmoidal function. \textit{Mathematics of Control, Signals, and Systems}, 5,\ 455-455.

[15]	Schuster, M.,\ \& Paliwal, K. K.\ (1997). Bidirectional recurrent neural networks. \textit{IEEE Transactions on Signal Processing}, 45,\ 2673-2681.

[16]	Shen, L.,\ Satta, G.,\ \& Joshi, A. K.\ (2007). Guided Learning for Bidirectional Sequence Classification. \textit{meeting of the association for computational linguistics},\ 760-767.

[17]	Reimers, N.,\ \& Gurevych, I.\ (2017). Optimal Hyperparameters for Deep LSTM-Networks for Sequence Labeling Tasks. \textit{arXiv: Computation and Language},\ 1707.06799.

[18]	Graves, A.,\ Mohamed, A.,\ \& Hinton, G. E. (2013). Speech recognition with deep recurrent neural networks. \textit{international conference on acoustics, speech, and signal processing},\ 6645-6649.

[19]	Srivastava, N., Hinton, G. E., Krizhevsky, A., Sutskever, I., \& Salakhutdinov, R. (2014). Dropout: a simple way to prevent neural networks from overfitting. \textit{Journal of Machine Learning Research}, 15,\ 1929-1958.

[20]	Maddison, C. J.,\ Mnih, A.,\ \& Teh, Y. W.\ (2016). The Concrete Distribution: A Continuous Relaxation of Discrete Random Variables. \textit{arXiv: Learning},\ 1611.00712.

[21]	Bengio, Y.,\ Leonard, N.,\ \& Courville, A. C.\ (2013). Estimating or Propagating Gradients Through Stochastic Neurons for Conditional Computation. \textit{arXiv: Learning},\ 1308.3432.

[22]	Han, S.,\ Pool, J.,\ Tran, J.,\ \& Dally, W. J.\ (2015). Learning both weights and connections for efficient neural networks. \textit{neural information processing systems},\ 1135-1143.

[23]	Huang, Z.,\ Xu, W.,\ \& Yu, K.\ (2015). Bidirectional LSTM-CRF Models for Sequence Tagging. \textit{arXiv: Computation and Language},\ 1508.01991.

[24]	Courbariaux, M.,\ Bengio, Y.,\ \& David, J.\ (2014). Training deep neural networks with low precision multiplications. \textit{arXiv: Learning},\ 1412.7024.

[25]	Jang, E.,\ Gu, S.,\ \& Poole, B.\ (2017). Categorical Reparameterization with Gumbel-Softmax. \textit{arXiv: Learning},\ 1611.01144.

[26]	Komninos, A.,\ \& Manandhar, S.\ (2016). Dependency Based Embeddings for Sentence Classification Tasks. \textit{north american chapter of the association for computational linguistics,}\ 1490-1500.

[27]   Zhang, Y., Chen, G., Yu, D., Yaco, K., Khudanpur, S.,\ \& Glass, J. R.\ (2016). Highway long short-term memory RNNS for distant speech recognition. \textit{international conference on acoustics, speech, and signal processing},\  5755-5759.

[28]   Ordonez, F. J.,\ \& Roggen, D.\ (2016). Deep Convolutional and LSTM Recurrent Neural Networks for Multimodal Wearable Activity Recognition. \textit{Sensors},\ 16(1).

%
%
%

\end{document}